\documentclass[journal,twocolumn,web]{article}
\usepackage{authblk}
\usepackage{framed,multirow}
\usepackage{amssymb}
\usepackage{latexsym}
\usepackage{amsmath}
\usepackage{url}
\usepackage{xcolor}
\usepackage{hyperref}
\usepackage{graphicx}

\providecommand{\keywords}[1]{\textbf{\textit{\\Keywords---}} #1}
\date{}

\renewcommand\footnotemark{}

\begin{document}

\title{A Quantitative Evaluation of Dense 3D Reconstruction of Sinus Anatomy from Monocular Endoscopic Video}

\author{
Jan~Emily~Mangulabnan, Roger~D.~Soberanis-Mukul, Timo~Teufel, Isabela~Hernández, Jonas~Winter, Manish~Sahu, Jose~L.~Porras, S. Swaroop~Vedula, Masaru~Ishii, Gregory~Hager, Russell~H.~Taylor and Mathias~Unberath
\thanks{``This work was funded in part by Johns Hopkins University internal funds and in part by NIH R01EB030511.
The content is solely the responsibility of the authors and does not necessarily represent the official views of the National Institutes of Health.'' }
\thanks{J.E. Mangulabnan (email: jmangul1@jhu.edu), R.D. Soberanis-Mukul (email: rsobera1@jhu.edu), T. Teufel (email: tteufel1@jh.edu), M. Sahu (email: msahu5@jhu.edu), S. S. Vedula (email: swaroop@jhu.edu), and G. Hager (email: hager@jhu.edu) are with Johns Hopkins University, Baltimore.}
\thanks{I. Hernández (email: i.hernandez@uniandes.edu.co) and J. Winter (email: jonas.winter@fau.de) were with Johns Hopkins University, Baltimore, when they participated in the project.}
\thanks{J.L. Porras (email: jporras1@jhmi.edu) and M. Ishii (email: mishii3@jhmi.edu) are with Johns Hopkins Medical Institutions, Baltimore.}
\thanks{R.H. Taylor (email: rht@jhu.edu) and M. Unberath (email: unberath@jhu.edu) are with Johns Hopkins University, Baltimore, and Johns Hopkins Medical Institutions, Baltimore.}
\thanks{J.E. Mangulabnan and R.D. Soberanis-Mukul are joint first authors. }
}

\maketitle

\begin{abstract}
Generating accurate 3D reconstructions from endoscopic video is a promising avenue for longitudinal radiation-free analysis of sinus anatomy and surgical outcomes. Several methods for monocular reconstruction have been proposed, yielding visually pleasant 3D anatomical structures by retrieving relative camera poses with structure-from-motion-type algorithms and fusion of monocular depth estimates. However, due to the complex properties of the underlying algorithms and endoscopic scenes, the reconstruction pipeline may perform poorly or fail unexpectedly. Further, acquiring medical data conveys additional challenges, presenting difficulties in quantitatively benchmarking these models, understanding failure cases, and identifying critical components that contribute to their precision. In this work, we perform a quantitative analysis of a self-supervised approach for sinus reconstruction using endoscopic sequences paired with optical tracking and high-resolution computed tomography acquired from nine ex-vivo specimens. Our results show that the generated reconstructions are in high agreement with the anatomy, yielding an average point-to-mesh error of 0.91\,mm between reconstructions and CT segmentations. However, in a point-to-point matching scenario, relevant for endoscope tracking and navigation, we found average target registration errors of 6.58\,mm. We identified that pose and depth estimation inaccuracies contribute equally to this error and that locally consistent sequences with shorter trajectories generate more accurate reconstructions. These results suggest that achieving global consistency between relative camera poses and estimated depths with the anatomy is essential. In doing so, we can ensure proper synergy between all components of the pipeline for improved reconstructions that will facilitate clinical application of this innovative technology.
\end{abstract}
\keywords Dense Reconstruction, Depth Fusion, Learning-based Descriptor, Sinus.

\section{Introduction}
\label{sec:introduction}

The nasal passage and paranasal sinuses give rise to several common yet difficult to treat medical conditions. For example, the nasal airways may come to be obstructed by nasal septal deviation which carries a reported prevalence of 70-80\,\% in the general population~\cite{clark2018nasal,moubayed2022evaluation,lee2021association}. When conservative management fails, the most frequently performed corrective surgery for nasal septal deviation in adults is septoplasty~\cite{manoukian1997recent}, which often improves quality of life, but its success is dependent on the pre-operative, patient-reported severity. In one study, 81\,\% of patients with severe nasal obstruction but only 31\,\% with mild nasal obstruction reported improvements 12 months after undergoing this procedure~\cite{pedersen2019prognostic}. These disappointing outcomes reflect a critical need to better understand the objective factors that may maximize the long-term success of surgery including, for example, optimal post-operative nasal geometry.

Similarly, sinusitis, a condition in which the mucosal lining of the paranasal sinuses becomes inflamed, is another opportunity for improving the objective means by which we both manage and monitor patients. Sinusitis is common in the United States, affecting one in six adults annually and it generates more than three million ambulatory care visits per year. Total direct health care expenditures attributed to sinusitis are estimated to be \$8.3 billion annually~\cite{bib:wyler2019,bib:ray1999,bib:bhattacharyya2011,bib:lethbridge2004}When medical treatment fails, sinusitis can be treated with functional endoscopic sinus surgery (FESS). The goals of FESS are to enlarge sinus ostia, restore aeration of the sinuses, improve mucociliary transport, and provide an optimal environment for topical therapy. Despite its widespread adoption, FESS remains a challenging procedure owing to anatomic variation and the range in severity of disease burden between patients. While surgeons can be assisted through use of intraoperative navigation to facilitate sinus dissection, it is limited in its ability to provide surgeons with a comprehensive cross-sectional understanding of each patient’s anatomy. 

Both nasal septal deviation and sinusitis are conditions reflecting a critical need for better methods of quantitative assessment of patient anatomy throughout the sinonasal airways. For example, longitudinal, quantitative assessments of nasal airway anatomy after septoplasty could facilitate an understanding of the relationships between anatomical structure, perceived symptoms, surgical outcomes, and changes in these relationships over time. Currently, computed tomography (CT) is the gold standard for acquiring accurate 3D representations of sinonasal anatomy making it a critical imaging modality for diagnosis and pre-operative planning. However, CT can be costly and exposes patients to ionizing radiation, thus limiting its use in large-scale longitudinal and cross-sectional analysis.

In this regard, endoscopy is a promising alternative for cross-sectional and longitudinal monitoring of long-term surgical outcomes as it is routinely employed in outpatient and clinical settings for qualitative assessment of sinonasal anatomy, is relatively inexpensive, and does not expose the patient to ionizing radiation. Endoscopy’s potential utility is not limited to longitudinal evaluation. For example, during endoscopic endonasal skull base surgery for repair of a CSF leak or resection of a tumor, the ability to generate a cross-sectional anatomical reconstruction could illuminate the relationship between a target lesion and critical anatomy. However, to enable cross-sectional and longitudinal quantitative comparisons of sinonasal airway anatomy, precise image-based reconstruction techniques with CT agreement are needed.  
This has motivated the development of methods for surface reconstruction from endoscopic sequences, including structure from motion-based methods~\cite{bib:phan2019,bib:widya2019,bib:qiu2018,bib:leonardimagebased2016,bib:reiter2016,bib:leonardevaluation2016,bib:hernandez2016}, Poisson reconstruction~\cite{bib:kazhdan2006}, Shape-from-Shading combined with fusion techniques~\cite{bib:turan2018,bib:tokgozoglu2012,bib:karargyris2011}, SLAM-based methods~\cite{bib:lamarca2019,bib:song2018,bib:ma2019,bib:liuslam2022}, tissue deformation models~\cite{bib:zhao2016,bib:lamarca2019,bib:song2018}, and learning-based methods~\cite{bib:ma2019,bib:chen2019}.

Surface reconstruction methods use volumetric fusion strategies, combining depth information obtained from sensors to estimate the 3D surface~\cite{bib:curlessvolumetric1996,bib:zachglobally2007}. 
For monocular video (the case for endoscopy), it has been of interest to employ deep neural networks for single image-based depth estimation. 
The prevailing image-based reconstruction method involves camera pose estimation using methods like structure from motion (SfM), which is combined with image-based depth estimation to generate the final surface~\cite{bib:liudreco2020,bib:ma2019,bib:reiter2016,bib:turan2018}. As ground-truth depth information is usually unavailable, recent works use self-supervised and unsupervised approaches, and these deep learning-based methods, like~\cite{bib:liudreco2020}, have shown promising results in generating three-dimensional sinus reconstructions.

\begin{figure}[t!]
    \centering
        \includegraphics[width=\columnwidth]{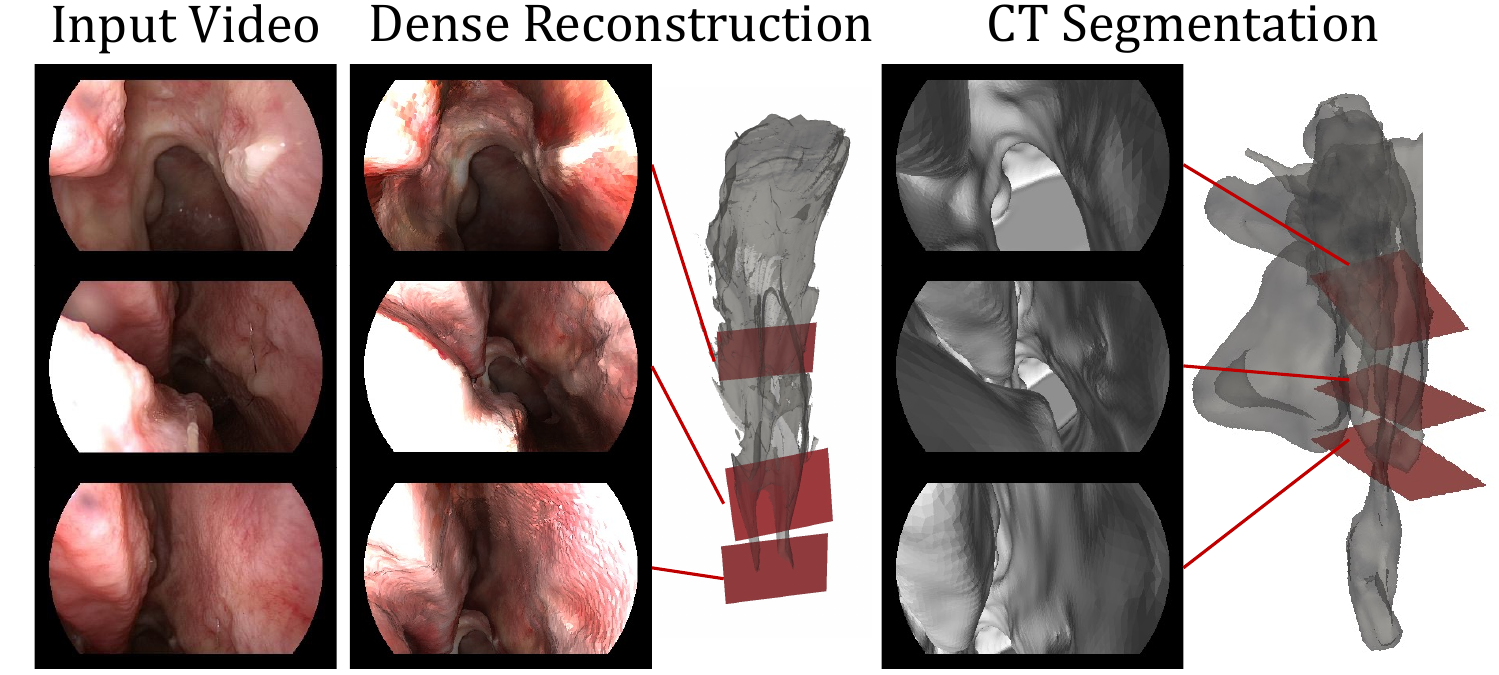}
        \caption{Results of the sinus dense reconstruction obtained with the method proposed in~\cite{bib:liudreco2020}. The columns: "Input Video" shows examples of endoscopic frames: "Dense Reconstruction" presents examples of the equivalent frames rendered from the reconstruction (showed in gray). The crossectional planes in the reconstruction indicate the camera position; "CT Segmentation" presents again renders of the same cameras but from the corresponding CT image.}
    \label{fig:dreco_results}
\end{figure}

In~\cite{bib:liudreco2020}, we presented a dense reconstruction method capable of generating watertight, visually pleasant sinus reconstructions. This approach enhances the initial SfM reconstruction by replacing the conventional descriptors with a learning-based descriptor~\cite{bib:liuextremely2020}. The SfM results are then used as a training signal for the depth estimator \cite{bib:liudepth2019}, resulting in high quality reconstructions, as presented in Fig. \ref{fig:dreco_results}.

These methods are shown to generate surfaces that offer a high level of agreement with the anatomy, reaching submillimeter precision on clinically relevant parameters \cite{bib:liudreco2020}. However, in a point-to-point matching scenario, relevant to endoscope tracking and navigation, its performance has yet to be fully explored, and an in-depth quantitative analysis of the main model components (i.e., depth and pose estimations) is required to understand the advantages and limitations of this and similar techniques toward practical clinical usage.

Unfortunately, such quantitative evaluation  
is difficult in practice, given the necessary data does not directly emerge from routine practice. 
In the medical setting, it is difficult to adapt clinically standard hardware like the endoscope camera to obtain depth and pose measurements, limiting the availability of adequate datasets for benchmarking anatomical reconstruction methods.
This lack of data makes it complicated to isolate and identify the critical components that contribute to the precision of the obtained 3D surfaces, such as inaccuracies in camera pose or monocular depth estimation, and hence a complete understanding of the influence of these components has not yet been fully addressed.

We contribute towards the quantitative assessment of sinus reconstruction and provide a rigorous evaluation of the reconstructions generated based on the work in~\cite{bib:liudreco2020} on an in-house dataset obtained from ex-vivo specimens. Our dataset includes endoscopic sequences, optical tracking data, and CT images, providing adequate ground-truth information for such assessment. 
Because this reconstruction pipeline involves components routinely employed in multiple monocular video-3D reconstruction frameworks, including~\cite{bib:ma2019,bib:reiter2016,bib:phan2019,bib:widya2019,bib:turan2018}, our work contributes important perspectives on the influence of individual components employed in the generation of the volumetric reconstruction. 

The rest of the paper is organized as follows: Section \ref{sec:method} first introduces the dense reconstruction strategy. We then present the details of the data collection method, dataset, and evaluation framework in section \ref{sec:eval_framework}. The quantitative analysis of the reconstructions is presented and discussed in Section \ref{sec:exp_res}. Finally, we conclude the paper in Section \ref{sec:conclusion}.

\section{Sinonasal Airway Reconstruction} \label{sec:method}

\begin{figure}[t!]
    \centering
        \includegraphics[width=\columnwidth]{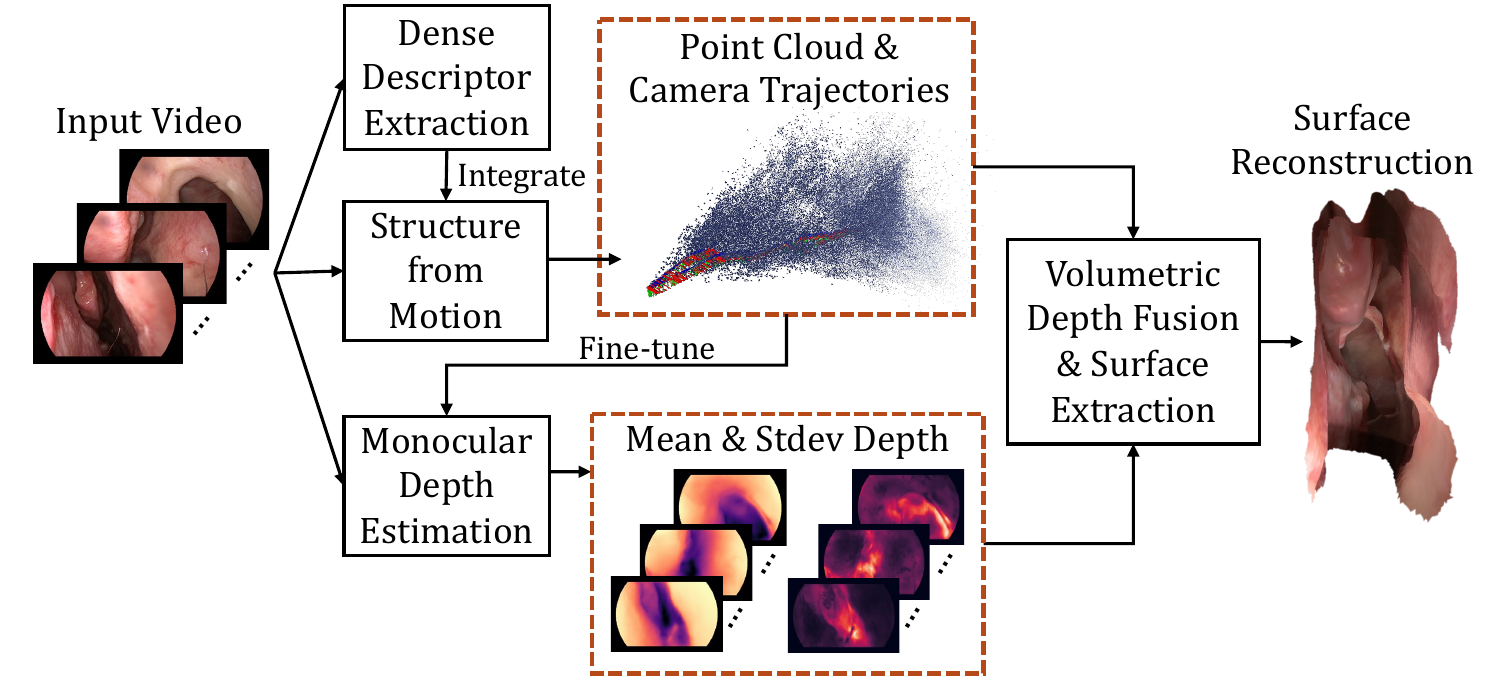}
        \caption{The dense reconstruction pipeline presented in \cite{bib:liudreco2020}. The process follows three main stages. First, SfM is run using a learning-based descriptor to obtain camera trajectories and dense point cloud. Then this point cloud is employed as supervisory signal to fine-tune the depth estimator. Finally, depth predictions are generated for each frame and fused to generate the final surface.}
    \label{fig:pipeline}
\end{figure}

Our previously proposed pipeline~\cite{bib:liudreco2020} aims to reconstruct a watertight sinonasal airway anatomical surface from unlabeled endoscopic videos. The pipeline has three main components shown in Fig. \ref{fig:pipeline}: 1) SfM based on learning-based dense descriptors, 2) monocular depth estimation, and 3) volumetric depth fusion and surface extraction. 

\subsection{Structure from Motion with Dense Descriptor} \label{sec:dense_descriptor}

Structure from motion \cite{bib:schonbergersfm2016} recovers camera pose and scene 3D structure from a sequence of images obtained from calibrated or uncalibrated cameras. This algorithm produces a sparse 3D reconstruction of scene points and estimated camera trajectory for each frame in the video sequence. SfM employs 2D correspondences usually generated with keypoints represented by hand-crafted descriptors like the scale invariant feature transform (SIFT) proposed by Lowe~\cite{bib:lowesift2004}. However, hand-crafted descriptors employed in anatomical surfaces present short tracking lengths and present problems with the repetitive textures present in the anatomical scenes, which hinder the extraction of adequate correspondences.

In order to resolve this problem and improve the density of the sparse reconstruction, recent methods employ learning-based descriptors as a replacement for hand-crafted descriptors during the point correspondence phase of SfM. This approach has been shown to yield better results and can significantly enhance the overall performance of the system. We refer to this process as Dense Descriptor Extraction. We define a particular pixel $i$ of frame $s$ as $\mathbf{x}_{i}^s$. Note that $\mathbf{x}_{i}^s$ is a two-dimensional vector representing the pixel's column and row position. We use the notation $F_{i}^s=  F(\mathbf{x}_{i}^s)$ to define the computed feature descriptor at position $\mathbf{x}_{i}^s$. $F(\mathbf{x}_{i}^s)$ is trained in a self-supervised approach using a set of frames from an endoscopic sequence. Following~\cite{bib:liuextremely2020}, the self-supervised training signal comes from a sparse SfM reconstruction with feature matches computed using a hand-crafted descriptor such as SIFT~\cite{bib:lowesift2004}.  

In the experiments presented in this paper, the Dense Descriptor Extraction step employs a pre-trained descriptor $F(\mathbf{x})$. The pre-trained descriptor is employed to generate dense correspondences between the frames of the endoscopic sequence. Overall, the process to obtain the correspondences (feature matching) uses a keypoint detector to generate query 2D locations in the source frame. A response map is generated for each query location $\mathbf{x}_{i}^q$ in the query image by comparing its feature descriptor $F_{i}^q$ with the pixel-wise feature descriptors in the target image $F_{j}^t$, $j=1,\dots N$, with $N$ the total number of pixels in the target image $t$. To achieve subpixel matching accuracy, we further apply bicubic interpolation to the response map and use the position with the maximum response as the matching location~\cite{bib:liuextremely2020}. 

This approach broadly increases the number of correspondences generated in the sequence. When these correspondences are employed inside SfM, the density of the sparse reconstruction generated also improves. 
The dense descriptor and corresponding SfM results are then employed to fine-tune the depth estimation network. 

\subsection{Dense Depth Estimation} \label{sec:depth_estimator}

We train the depth estimation model with the results of the dense descriptor SfM by employing a self-supervised approach~\cite{bib:liudepth2019}. 
The 3D structure and correspondences obtained with SfM are used towards sparse depth guidance for flow (from depth) consistency between frames that is complemented with inter-frame dense depth consistency. Overall, the objective function for the dense estimation is defined as follows~\cite{bib:liudepth2019}:
\begin{equation}
    \mathcal{L}(j, k) = \lambda_1\mathcal{L}_f(j, k) + \lambda_2\mathcal{L}_d(j,k)
\end{equation}
where $\mathcal{L}_f$ and $\mathcal{L}_d$ respectively define the flow and depth consistency losses between the frames $j$ and $k$, and $\lambda_1$, $\lambda_2$ are balancing parameters. The flow loss aims to minimize the difference between dense flow maps (computed from depth estimation) and sparse depth maps (obtained from the SfM 3D structure) to generate depth maps that agree with the sparse SfM reconstruction. The depth consistency loss enforces geometric constraints between independently predicted depth maps for inter-frame agreement.

Based on improvements to the depth model in~\cite{bib:liudreco2020}, we additionally consider probabilistic depth estimates, as poorly illuminated areas may not provide sufficient information to enable precise depth estimation. To this end, depth is modeled as a pixel-wise independent Gaussian distribution represented by its mean $\mu$ and standard deviation $\sigma$. Depth consistency is enforced by minimizing the log-likelihood loss of the mean dense depth prediction $\hat{\mu}_{j, k}$ of frame $j$ warped to the reference view of frame $k$ with respect to the mean depth $\mu_k$ of frame $k$. This loss is defined by Eq. \ref{eq:dense_depth_consistency}.
\begin{equation}\label{eq:dense_depth_consistency}
    \mathcal{L}_{dns}(j,k) = \log{\sigma^2_k} + \frac{(\hat{\mu}_{j, k} - \mu_k)^2}{2\sigma^2_k}
\end{equation}

Furthermore, we included a related training objective that maximizes the joint probability of the training data from SfM given the predicted depth distribution. This loss function is formulated similarly to Eq. \ref{eq:dense_depth_consistency}, with the difference that it is defined over the sparse depth estimation obtained from the SfM results. The sparse probabilistic depth consistency is defined in Eq.  
\ref{eq:sparse_depth_consistency} as: 
\begin{equation}\label{eq:sparse_depth_consistency}
    \mathcal{L}_{sps}(j,k) = M_k\log{\sigma^2_k} + M_k\frac{(z_k - \mu_k)^2}{2\sigma^2_k}
\end{equation}
where $z_k$ is the sparse depth estimation defined by the $z$ components of the SfM 3D points that project onto the frame $k$ and $M_k$ is the corresponding sparse map of frame $k$ which defines a weighted factor for each $z_k$ based on the number of frames employed by SfM to triangulate the 3D points that generate $z_k$~\cite{bib:liudepth2019}.

We also add an appearance consistency loss~\cite{bib:liudreco2020}, which is commonly used in self-supervised depth estimation for natural scenes where photometric constancy assumptions are reasonable~\cite{bib:zhou2017}. This assumption is invalid in endoscopy and cannot be used for additional self-supervision. However, the pixel-wise descriptor map from the Dense Descriptor Extraction module is naturally illumination-invariant and provides a dense signal. It can thus be incorporated into appearance consistency, where appearance is defined in terms of the dense descriptor (Eq. \ref{eq:appearance_consistency}).
\begin{equation}\label{eq:appearance_consistency}
    \mathcal{L}_{apc}(j,k) = ||\hat{F}^{j,k} - F^k||^2
\end{equation}
The definition of appearance consistency loss in Eq. \ref{eq:appearance_consistency} compares the features maps $F^k$ of the frame $k$ with the features $\hat{F}^{j,k}$ of frame $j$ warped to the positions of the frame $k$. 

The architecture of the depth estimator follows~\cite{bib:liudepth2019} with the exception that the last layer is replaced by two small heads to generate both a mean depth map and a standard deviation depth map. The overall training objective is defined by the combination of all the loss functions defined in Eq. \ref{eq:dense_depth_consistency}, \ref{eq:sparse_depth_consistency}, and \ref{eq:appearance_consistency} in addition to the flow consistency $\mathcal{L}_{f}$ defined in~\cite{bib:liudepth2019}: 
\begin{equation}\label{eq:depth_loss}
    \mathcal{L} = \lambda_1\mathcal{L}_{dns} + \lambda_2\mathcal{L}_{sps} + \lambda_3\mathcal{L}_{apc} + \lambda_4\mathcal{L}_{f}
\end{equation}
Based on the sparse supervision from SfM, together with the dense constraints of geometric and appearance consistency, the network learns to predict accurate dense depth maps with uncertainty estimates for all frames, which are fused to form a surface reconstruction.

\subsection{Depth Fusion} \label{sec:depth_fusion}

The depth fusion strategy uses a set of depth measurements to estimate the truncated signed distance function (TSDF) that is then employed to build a volumetric representation of the surface. 
As previously introduced, this information comes from depth prediction models for monocular endoscopic videos.
For each 3D point $\mathbf{X}$, depth measurements obtained with our depth estimator are propagated to a 3D volume using ray-casting from the corresponding camera pose $i$ to estimate the signed distance function $d^{*}_i(\mathbf{X})$. The truncated distance~\cite{bib:curlessvolumetric1996,bib:zachglobally2007} is then defined as:
\begin{equation}
    d_i(\mathbf{X}) = \min(1, \frac{d^{*}_i(\mathbf{X})}{\sigma})
\end{equation}
Note that the corresponding uncertainty estimates $\sigma$ determine the slope of the truncated signed distance function for each ray. All cameras that have a visible projection of $\mathbf{X}$ are integrated employing~\cite{bib:curlessvolumetric1996,bib:zachglobally2007}:
\begin{equation} \label{eq:depth_fusion}
    D(\mathbf{X}) = \frac{\sum w_i(\mathbf{X})d_i(\mathbf{X})}{\sum w_i(\mathbf{X})}
\end{equation}

The baseline reconstructions were created using equal weighting ($w_i(\mathbf{X}) = 1$) for all the distance estimations. To fuse all information correctly, the camera poses estimated from SfM are used to propagate the corresponding depth estimates and color information to the 3D volume. Finally, the Marching Cubes method~\cite{bib:lorensenmarchingcubes1987} is used to extract a watertight triangle mesh surface from the 3D volume.

\section{Evaluation Framework}\label{sec:eval_framework}
We perform an exhaustive analysis of the predicted depth, camera pose, and volumetric fusion components of our process. In order to establish a reference to compare the obtained reconstructions, we implement a data collection process and define a comparison procedure that allows a direct evaluation against CT imaging. In this section, we describe the details of the data collection and evaluation procedures. 

\subsection{Evaluation Dataset} \label{sec:datacollection_protocol}

\begin{figure}[t!]
    \centerline{\includegraphics[width=\columnwidth]{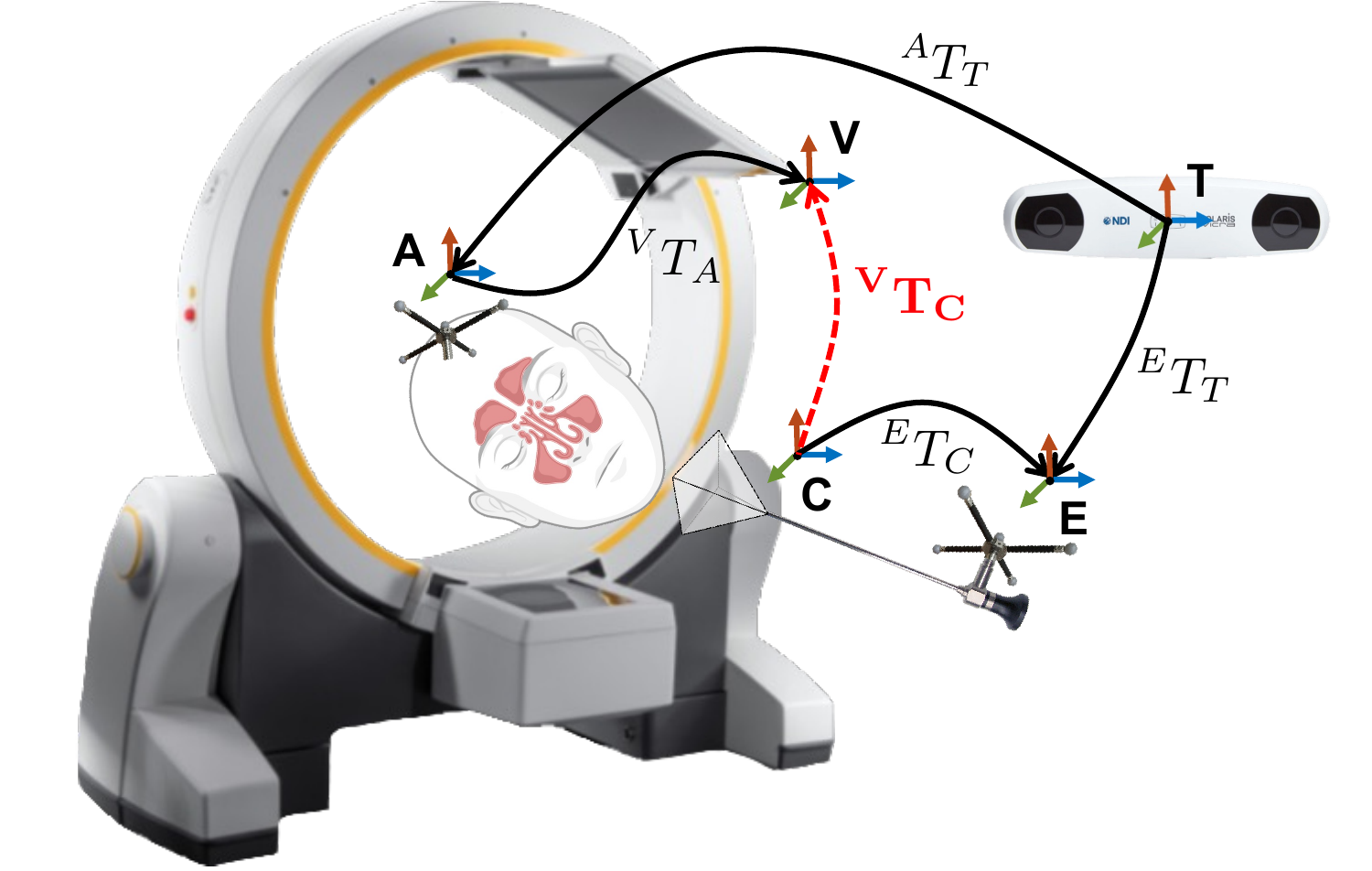}}
    \caption{Configuration of the data collection protocol employed to obtain a data set of CT scans paired with video and pose information. The optical tracker provides timestamped information of the reflective marker geometries placed in the endoscope and the anatomy. Additionally, the timestamped video information is also recorded.}
    \label{fig:data_collection}
\end{figure}

We implemented a data collection protocol allowing us to acquire CT data and optical tracking information during a simulated pre-operative sinus exploration on ex-vivo subjects. The scoping was performed by an experienced surgeon using a rigid endoscope connected to a Storz Image1 HD camera (Karl Storz SE \& Co. KG, Tuttlingen, Germany).
Rigid geometries equipped with retro-reflective spheres are attached to both the camera and anatomical specimen, allowing movement tracking relative to the optical tracker. Tracking information was acquired with an NDI Polaris Hybrid Position Sensor (Northern Digital Inc., Waterloo, Canada).
Prior to the study, we performed a checkerboard-based hand-eye calibration \cite{bib:strobloptimal_handeye2006} to obtain the camera position relative to the camera marker geometry. Additionally, we perform a checkerboard camera calibration to obtain the endoscope's intrinsic parameters.
We obtained a CT scan of the anatomy after the exploration with a Brainlab LoopX scanner (Brainlab, Munich, Germany), ensuring that both sinus anatomy and rigid marker geometry are visible in the CT scan for tracker-to-CT registration. All the information related to the optical tracker and the endoscope video signal was recorded using a ROS-based platform. The recordings include time stamps allowing pair-match of the Polaris poses with each frame in the endoscopic sequences.  

We manually segmented the sinus anatomy and the reflective spheres in the obtained CT scans before our analysis. The segmentation of the spheres was registered to the tracked anatomical coordinate frame to define the camera poses in CT space ($^VT_C$).This transformation chain is shown in Fig. \ref{fig:data_collection} and described as:
\begin{equation} \label{eq:transformation_chain}
    ^VT_C =\:^VT_A\,^AT_T\,(^ET_T)^{-1}\,^ET_C
\end{equation}
where $^ET_C$ is the hand-eye calibration, $^AT_T$ and $^ET_T$ are the tracked anatomy and endoscope frames, respectively, and $^VT_A$ is the registration transformation.
We collected data on eight cadaveric specimens sourced from the Maryland Anatomy Board. In cases where anatomical conditions allowed, both nostrils were inspected. In total, the endoscopic videos were used to generate nine dense reconstructions that could be analyzed with respect to corresponding CT and Polaris data.

\subsection{Anatomy and Reconstruction Coordinate Frames} \label{sec:frames}

The analysis occurs between the CT volume coordinate frame $V$ and the reconstruction coordinate frame $R$. We also employ two surfaces defined in each coordinate frame: the CT segmentation surface $S_A$ and the reconstructed surface $S_R$. In addition to the surfaces, we also consider two sets of camera poses  $^{V}T_{C}$ given by the optical tracker and $^{R}T_{C}$ obtained during SfM, which represent the camera-to-world transformation of the same images but with respect to their respective anatomical volume and reconstruction spaces. This section describes the process to obtain the registered reconstruction $S_{VR}$ to the anatomy frame, where the subscript ${VR}$ denotes the surface $S_{R}$ represented in the coordinate system of $V$ (a similar convention will be employed with other transformations when convenient).

\subsection{Reconstruction-CT Registration} \label{sec:pose_registration}

\begin{figure}[t]
    \centerline{\includegraphics[width=\columnwidth]{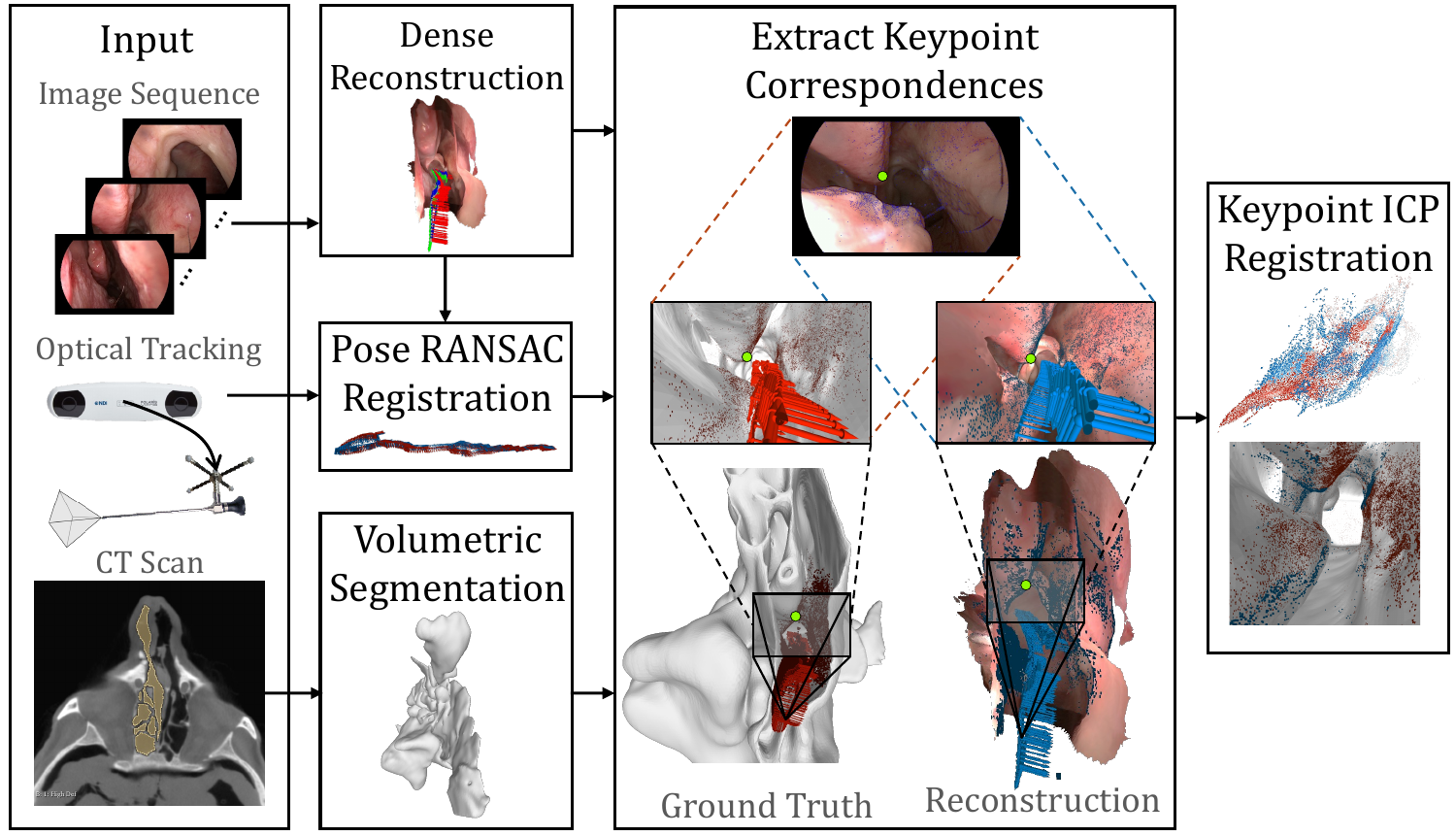}}
    \caption{Overall pipeline of the registration framework. We first employ the pose information from the SfM estimation and the optical tracker to establish an initial pose-based registration that is further refined with ICP between corresponding keypoints obtained by re-projecting the 2D locations to the corresponding reconstruction and CT segmentation, using depth rendered from their corresponding camera pose set.}
    \label{fig:reg_pipeline}
\end{figure}

Initial registration was achieved using the one-to-one correspondence between the camera poses $^{V}T_{C}$ and $^{R}T_{C}$ for each frame.
Considering the scale ambiguity present in monocular-based reconstruction methods, we first use the relative distances between consecutive cameras to scale $S_V$ to $S_R$. We then use a linear solver to find the transformation $^{V}T_{R}$ between the reconstruction and anatomy volume coordinate frames based on the poses. We additionally apply a RANSAC algorithm to account for potential outliers between the SfM camera estimations and the optical tracking poses. Finally, we apply the transformation to the camera poses and reconstruction to obtain the initial registered reconstruction:
\begin{equation}
    S_{VR} = ^{V}T_{R} \cdot S_R
\end{equation}
and corresponding camera poses:
\begin{equation}
    ^VT_{RC} = ^{V}T_{R} \cdot ^{R}T_{C}
\end{equation}
where ${RC}$ indicates that the estimated camera poses were defined in the anatomy volume space. 

Considering that the pose registration is based only on camera centers and our main goal is to evaluate the congruence between two 3D surfaces, we apply a complementary point-based registration between the segmentation and reconstruction surfaces. 
After obtaining the initial pose-based registration $^{V}T_{R}$, we further refine the transformation by applying an ICP algorithm over a set of corresponding 3D points defined between the surfaces. 
Our objective is to separate errors due to camera pose estimation and analyze the results when the registration is inherent to both surfaces. In addition, registering with ICP increases the number of 3D correspondences employed in the registration beyond the number of cameras, leading to a more precise registration.

To generate these correspondences, we consider that each camera pose is self-consistent with its respective surface, either because they were acquired employing optical tracking or co-optimized along with the surface during the reconstruction process. 
Under this assumption, we employ the surfaces and camera poses to render the per-camera depth maps $D_{VV}(\mathbf{x})$ and $D_{VR}(\mathbf{x})$ for each image employed to generate the reconstruction. Note that $D_{VR}(\mathbf{x})$ was computed after applying the initial pose-base registration to the reconstruction elements. Given that the set of images is common to both surfaces, we create a set of 2D points in a given image and use the rendered depth to reproject the point into both surfaces to obtain their corresponding 3D projection. For the anatomy CT segmentation, this relationship is defined by equation \ref{eq:ct_reprojection}:
\begin{equation}\label{eq:ct_reprojection}
    \mathbf{X}_{VV} = {^V}T_C\cdot D_{VV}(\mathbf{x}) \cdot K_V^{-1} \cdot \mathbf{x} 
\end{equation}
where $K_V$ refers to the camera intrinsic parameters estimated by calibrating the endoscope camera during the data collection. Using a similar approach, we can compute the corresponding point in the reconstructed surface, as indicated by equation \ref{eq:dreco_reprojection}:
\begin{equation}\label{eq:dreco_reprojection}
    \mathbf{X}_{VR} = (^{V}T_{R}{^R}T_C)\cdot D_{VR}(\mathbf{x}) \cdot K_R^{-1} \cdot \mathbf{x} 
\end{equation}

Note that $K_r$ represents the estimated SfM camera intrinsics in this case. Given the self-consistency of the poses and surfaces, these re-projections generate a set of correspondences between the anatomy and the reconstruction surfaces that we employ to fine-tune the initial registration leading to a refined transformation:
\begin{equation}
    ^{V}T_{VR} = ICP(\mathbf{X}^i_{VV}, \mathbf{X}^i_{VR}), i=1\dots N
\end{equation}
with N representing the number of reprojected points. We sample points using the SIFT keypoint detector from all the images of the sequence to generate the correspondences. By applying this transformation to the constructed surface, we obtain the refined registered surface $S^{'}_{ar}$ defined as
\begin{equation}
    S^{'}_{VR} = ^{V}T_{VR} \cdot S_{VR}
\end{equation}

During our experiments, we compare the ICP-refined registration of the surface reconstruction with the segmented anatomical CT to analyze the agreement of the generated surfaces and camera poses.  

\section{Experiments and Results} \label{sec:exp_res}

We start in Section \ref{sec:ct_agreement} with the evaluation of the baseline method that employs the vanilla configuration of the pipeline described in Section \ref{sec:method}. Then, in order to understand the influence of the different components on the accuracy of the reconstruction, we analyze the influence of uncertain depth predictions during the depth fusion stage (Section \ref{sec:depth_fus_comp}). An assessment of the influence of the depth and pose predictions is presented in Section \ref{sec:ct_depth_fusion}. Finally, we explore the performance of the reconstructions under short-track trajectories in Section \ref{sec:local_reconstruction}. 

\textbf{Evaluation Metrics} In all the experiments, we report the average point-to-mesh and target registration error (TRE) between the CT segmentation and the reconstructed anatomy. Point-to-mesh is computed directly between the CT segmentation and the reconstruction. To compute the TRE, we sampled a grid-like set of keypoints for each frame that then are reprojected onto the CT segmentation and reconstruction surfaces using the reprojection approach described in \ref{sec:pose_registration} (see Eq. \ref{eq:ct_reprojection}, and Eq. \ref{eq:dreco_reprojection}).

\subsection{Reconstruction-CT Agreement} 
\label{sec:ct_agreement}
We aim to evaluate the reconstruction with respect to the CT, as it is the gold standard method of obtaining the ground-truth 3D anatomical structure. We manually generated the sinus segmentation of the CTs corresponding to each of the nine endoscopic sequences. As our dataset also includes ground-truth camera trajectories, we use the segmented CT and reconstructed 3D surfaces together with  camera intrinsics and poses to establish 3D point correspondences between the reconstruction and segmentation independent of each other to report the target registration error. This allowed us to analyze how specific regions and structures of the sinus anatomy are mapped to the 3D mesh. The obtained results are reported in Table \ref{tab:base_results}.
\begin{table}[h!]
\centering
\caption{Average and standard deviations of the Target Registration error (TRE) and Point-to-Mesh Distance between the reconstructed anatomy and the CT segmentation.}
\label{tab:base_results}
\resizebox{\columnwidth}{!}{
\begin{tabular}{ccc}
\hline
\textbf{Reconstruction} & \textbf{TRE (mm)} & \textbf{Point-to-Mesh (mm)} \\ \hline
Baseline & 6.58 $\pm$~2.5 & 0.91 $\pm$~0.4 \\ 
\hline
\end{tabular}%
}
\end{table}

\begin{figure}[t] 
    \centerline{\includegraphics[width=0.99\columnwidth]{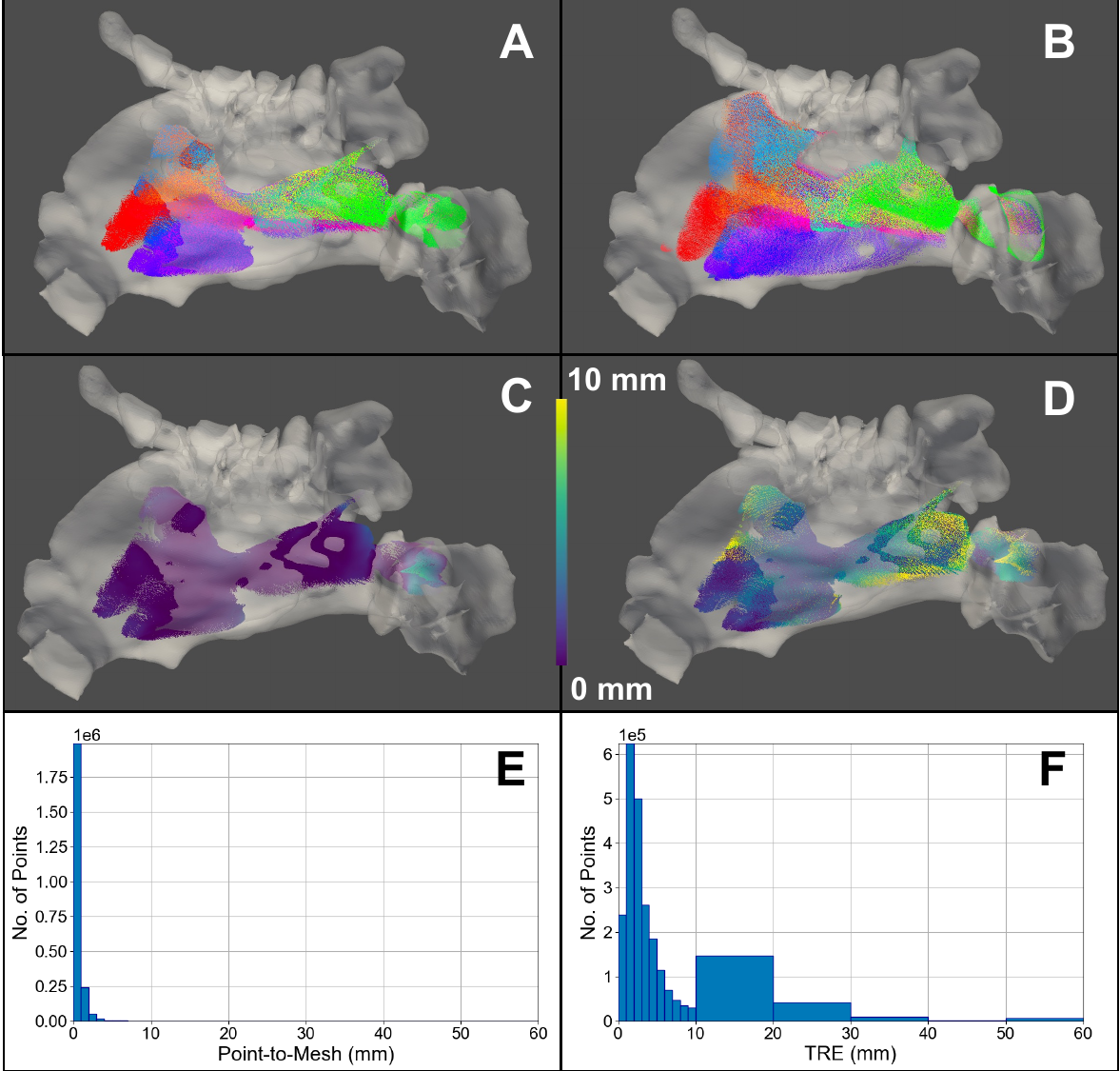}}
    \caption{Top Row: Spatial distribution of the color-coded correspondences employed for TRE computation for: (A) the dense reconstruction and (B) the CT segmentation. The same color indicates a correspondence, and the keypoints are presented in contrast with the full CT segmentation (indicated in gray). Middle Row: Spatial distribution of errors of the reconstruction's keypoints for: (C) the point-to-mesh distance, and (D) the TRE with respect to the CT segmentation (in gray). Bottom Row: Histograms of point-to-mesh (E) and TRE (F) distributions.}
    \label{fig:tre_correspondences}
\end{figure}

Results show a strong point-to-mesh agreement with millimeter precision at 0.91\,mm, suggesting that the reconstruction is geometrically similar to the CT segmentation. However, the TRE of 6.58\,mm indicates that the anatomical structures present in the endoscopic images are not necessarily mapped correctly on the final 3D reconstruction. The correspondences used to compute the TRE are visualized in Fig. \ref{fig:tre_correspondences} A and B, which shows color-coded 3D point projections onto the reconstructed and CT meshes, respectively. While the general locations of the correspondences within the CT segmentation are comparable based on similar color mapping, there appears to be warping in the overall structure resulting in a high TRE. This indicates difficulties in maintaining consistency throughout the entire sinus reconstruction. 

As the dense reconstruction employs SfM results as the main self-supervisory signal, we can guarantee that the estimated poses and the generated point cloud are consistent with respect to the 2D projections in the images. However, the ambiguity present in the process may lead to locally consistent solutions that might not propagate to the global context. This is evident in Fig. \ref{fig:tre_correspondences} D, which shows low TRE (less than 5\,mm) in a localized region, but higher errors in others. Additionally, the TRE distribution (Fig. \ref{fig:tre_correspondences} F) indicate that the majority of the point correspondences are mapped correctly, further implying that the relative mapping of certain regions becomes distorted when reconstructing the anatomy.

Considering both the point-to-mesh and TRE results, the surface may appear similar to the ground-truth reference, but these points may be close to the sinus wall or other structures that are not the true corresponding points. In order to investigate potential sources of error that warp regions of the reconstruction (and thus contribute to the TRE), we analyzed the fusion method from multiple perspectives, including the influence of uncertainty from the predicted depth maps and the accuracy of the pose and depth predictions.

\subsection{Sensitivity of Depth Fusion to Uncertainty} 
\label{sec:depth_fus_comp}

An important part of the fusion strategy is the computation of the signed distance function, which depends on the estimated depth. To this end, we analyzed the sensitivity of the reconstruction to the potential presence of inconsistent depth predictions. In particular, we consider that noise may be present in the current predictions depending on the accuracy of the SfM results introducing errors in certain regions of the final reconstruction. Hence, we evaluate the generated reconstruction when: 1) the depth estimations are weighted based on uncertainty; 2) outliers are removed from the predicted maps; and 3) when removing large depth predictions from the depth maps (considering that surfaces close to the camera are better illuminated, and hence have more confidence estimations). Results for these experiments are shown in Table \ref{tab:fusion_results}.

\begin{table}[h]
\centering
\caption{Average and standard deviations of the Target Registration error (TRE) and Point-to-Mesh Distance between the CT segmentation and different ablations of the input depth for the volume fusion stage (TSDF estimation).}
\label{tab:depth_fusion}
\resizebox{\columnwidth}{!}{
\begin{tabular}{ccc}
\hline
\textbf{Reconstruction} & \textbf{TRE (mm)} & \textbf{Point-to-Mesh (mm)} \\ \hline
Baseline & 6.58 $\pm$~2.5 & 0.91 $\pm$~0.4 \\ 
\hline
Weighted & 6.47 $\pm$~2.4 & 0.87 $\pm$~0.3 \\ 
Outlier Removal & 6.98 $\pm$~2.3 & 0.90 $\pm$~0.3 \\ 
Large Depth Removal & 8.24 $\pm$~2.9 & 0.91 $\pm$~0.3\\ 
\hline
\end{tabular}%
}
\end{table}

\subsubsection{Uncertainty Weighting} 
\label{sec:weighting}

Due to the potential contribution of high uncertainty depths in the fusion step, we weighted the depth predictions based on confidence values to examine its effect on the final structure. Given that the depth estimation network generates standard deviation maps (representing uncertainty) corresponding with its predictions, we employed these values in Eq. \ref{eq:depth_fusion} to set the parameter $w_i(x)$ (originally weighted equally) as the inverse uncertainty at the pixel level.

Table \ref{tab:depth_fusion} - Weighted shows that weighting by uncertainty does not majorly change the mesh structure, resulting in only slight improvements of 0.11\,mm and 0.04\,mm in TRE and point-to-mesh, respectively. This may be due to depth consistency loss enforced during the training of the depth estimator, which reduces variability in the predictions. This means that even though some depth values might have high uncertainty, the predictions are optimized for a level of agreement with adjacent frames. 
It is also possible that high-uncertainty components are located near the borders of the 3D geometry of the anatomy (elements that generate edges in the image). In this sense, the camera movement during the integration step of the TSDF contributes to averaging these high uncertainty elements, even in the case when all depths are weighted with the same value ($w_i$ = 1), which can also lead to small differences between both generated reconstructions.

\subsubsection{Removal of Outliers} \label{sec:outlier_removal}

To further explore the impact of the uncertainty in the TSDF computation, we also evaluated the resulting reconstruction when these high uncertainty values are considered as outliers and not included at all in the depth fusion step. We removed the depth values with uncertainties that are beyond one standard deviation (68\,th percentile).

Results are presented in Table \ref{tab:depth_fusion} - Outlier Removal. Similar to the experiments with the weighting of the TSDF estimation (see Section \ref{sec:weighting}), differences between the standard pipeline and the reconstruction with the outliers removed are sub-millimetric. It is possible that this contributes to removing small noisy sections, but the contribution of the overall geometry is minor. Again, it is possible that given the concentration of the high uncertainty elements in the border regions of the anatomy, its contribution is already lowered even in the original pipeline. 

\subsubsection{Removal of Large Depths}
Prior modifications based on uncertainty were investigated based on the assumption that larger depths may be correlated with high uncertainty. Since endoscopic video is collected using a light source attached to the camera, surfaces further from the endoscope (i.e., deeper in the nasal cavity) are not well illuminated. As this may introduce noise in the depth fusion step when resolving depth values, we removed large depths beyond one standard deviation (68th percentile) to evaluate the contribution of these components in the generated reconstruction.
Table \ref{tab:depth_fusion} - Large Depth Removal shows the TRE and point-to-mesh distance obtained during this analysis. It appears that removing these elements has a negative impact on the TRE, meaning that large depths positively contribute to the final result, even though they could potentially be less reliable than short depths estimations.
In this regard, it is possible that improving the predicted depth maps would bring benefits to the reconstruction. 

\subsection{Pose and Depth Estimation Accuracy in Depth Fusion} \label{sec:ct_depth_fusion}

Depth and pose information are the main inputs to the estimation of the TSDF, and hence these components drive the reconstruction. Considering the availability of CT information and optical tracking in our in-house dataset, we ablate the depth and pose components that are employed in the approximation of the TSDF during the volumetric fusion.

\begin{table}[h]
\centering
\caption{Average and standard deviations of the Target Registration error (TRE) and Point-to-Mesh Distance between different ablations for the estimated trajectories and depth maps.}
\label{tab:fusion_results}
\resizebox{\columnwidth}{!}{
\begin{tabular}{ccc}
\hline
\textbf{Reconstruction} & \textbf{TRE (mm)} & \textbf{Point-to-Mesh (mm)} \\ \hline
Baseline & 6.58 $\pm$~2.5 & 0.91 $\pm$~0.4 \\ 
\hline
CT Depth - Polaris Poses & 0.65 $\pm$~0.3 & 0.08 $\pm$~0.1 \\
CT Depth - SfM Poses & 5.90 $\pm$~3.3 & 0.92 $\pm$~0.9 \\
Predicted Depth - Polaris Poses & 6.54 $\pm$~3.4 & 1.03 $\pm$~0.4 \\
\hline
\end{tabular}%
}
\end{table}

Using the optical tracker trajectories and the CT segmentation, it is possible to obtain reliable depth values by rendering the depth from the CT at the position of the tracked camera. We can consider this scenario the ideal input for the volumetric fusion method and the case where poses and depth are predicted with high accuracy. As shown in the results of Table \ref{tab:fusion_results} (CT Depth - Polaris Poses), the fusion of CT depths rendered at the Polaris poses does not introduce meaningful errors in the reconstructed mesh evident by sub-millimeter TRE and point-to-mesh distances. 

In the following experiments, we ablate both components, poses and depth estimations, by replacing the ideal inputs with the predictions from the depth estimator and the SfM poses independently. First, we utilized CT segmentation to render depths at the position of the camera trajectories given by the optical tracker, generating precise depth maps. However, in the volumetric fusion step, we employ the poses estimated by SfM (CT Depth - SfM Poses in Table \ref{tab:fusion_results}). This allowed us to investigate the contribution of the errors in pose estimation in the agreement of the reconstruction with the CT.
Results show nominal improvements in the obtained TRE compared with the baseline, while the point-to-mesh distance remains at a similar range. This suggests that even with accurate depth estimates, inaccurate camera poses impact the overall structure of the mesh.

\begin{figure}[t]
    \centerline{\includegraphics[width=0.99\columnwidth]{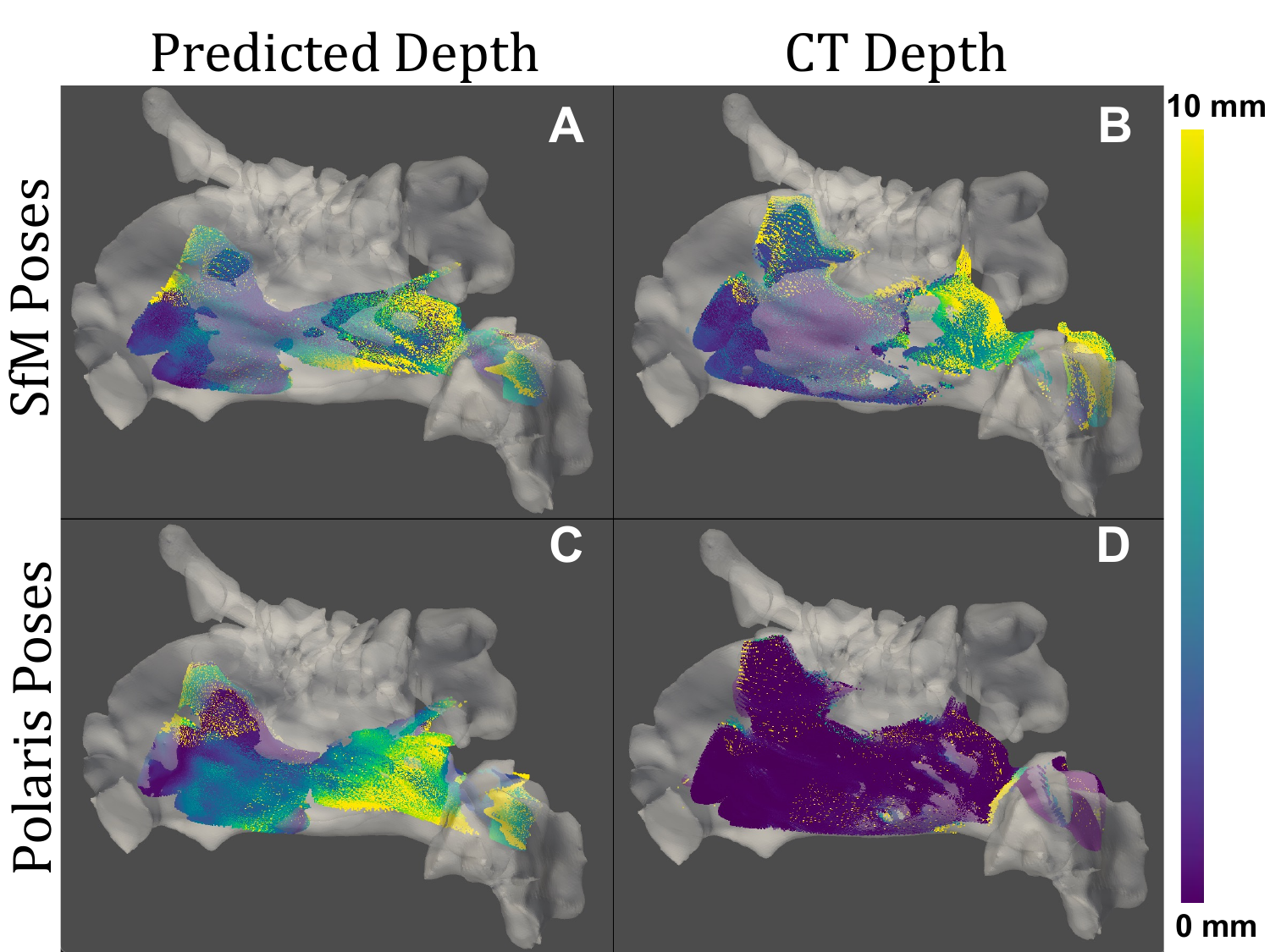}}
    \caption{Point TRE of (A) baseline reconstruction, which uses predicted depths and SfM poses, with reconstructions using (B) CT depths at SfM poses, (C) predicted depths at Polaris poses, and (D) CT depths at Polaris poses.}
    \label{fig:ct_depth}
\end{figure}

Furthermore, we maintained use of Polaris poses but substituted the CT-rendered depth with the predicted depth estimates from the model for each frame in the depth fusion method (Predicted Depth - Polaris Poses in Table \ref{tab:fusion_results}). The reconstructed mesh, in this case, also presented a similar TRE as the baseline and a higher point-to-mesh error. 
It is possible that the depth consistency enforced between frames generates a strong relationship between the poses employed in the training process and the predicted depth. From one perspective, as the results of Section \ref{sec:weighting} suggest, this has the benefit of reducing the effect of uncertain prediction, as the depth estimation between adjacent frames will retain a level of consistency. On the other hand, this strategy can be sensitive to the precision of the training signal given by the camera poses. As both components, poses, and depth, play a significant role in the volumetric fusion, it seems that uncertainties in the estimation of any of these components lead to inconsistencies in the generated reconstructions, indicating that motion and depth should not only be self-consistent but also co-optimized based on anatomical priors that ensure adequate constraints to reduce the variability of the solutions. 
Fig. \ref{fig:ct_depth} shows the spatial distribution of the TRE errors for all the reconstructions presented in Table \ref{tab:fusion_results}.

\begin{figure}[t]
    \centerline{\includegraphics[width=0.99\columnwidth]{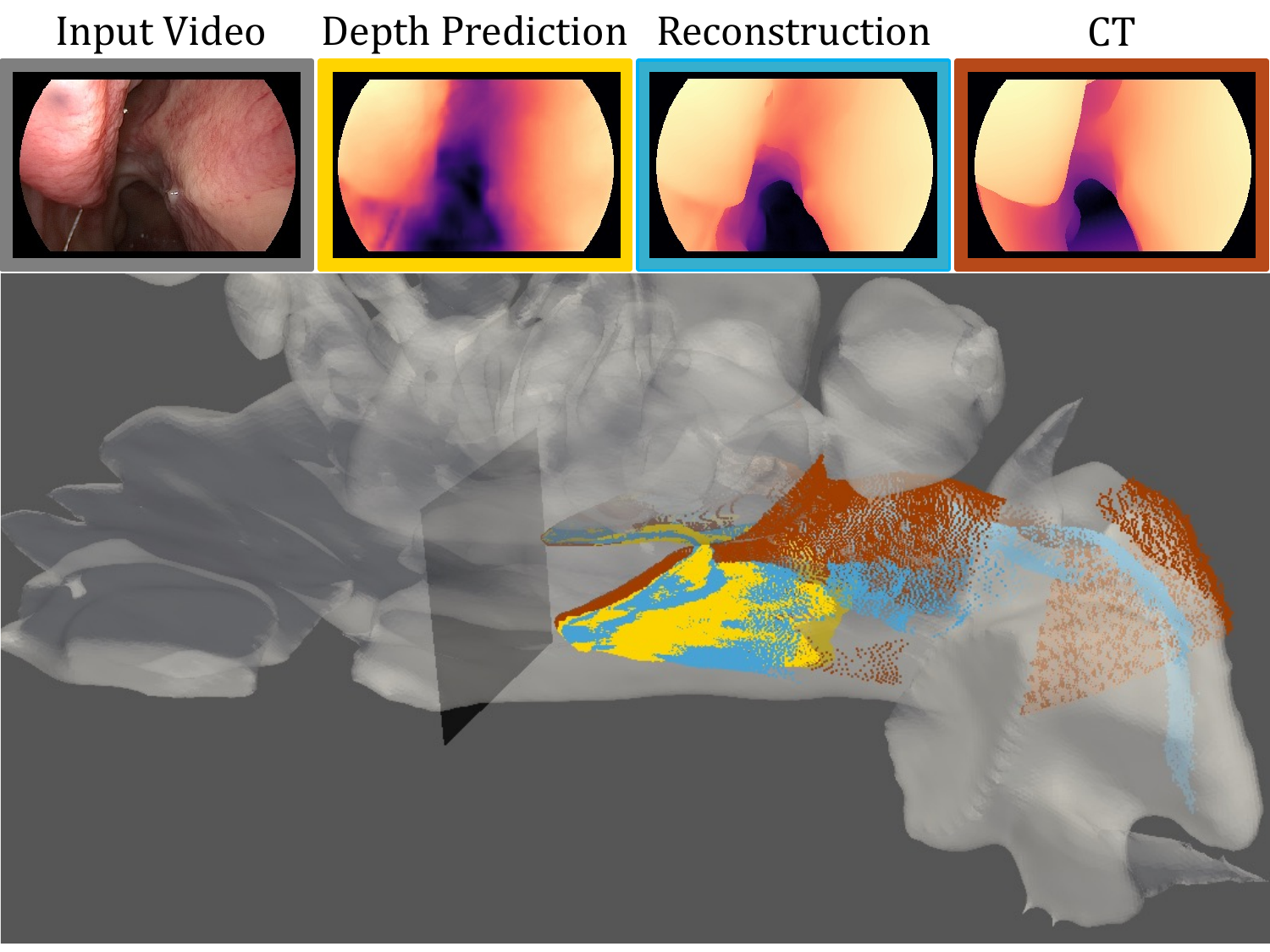}}
    \caption{Top: Visual comparison of a single input frame and depth maps from the model and rendered from the reconstruction and CT. Bottom: Comparison of 3D points projections from the camera frame (shadowed plane) using the predicted depth (yellow) and rendered depth from the reconstruction (blue) and CT segmentation (red).}
    \label{fig:depth_estimates}
\end{figure}

We further studied the error of the depth predictions by reprojecting the keypoints inside a given frame directly employing its corresponding depth prediction instead of rendered depth from the fused mesh. Then we compared these projections with respect to the CT. Note that this process is done at the (un-fused) frame level, and the average metrics across all available frames are reported. Results are shown in Table \ref{tab:depth_projections}. A visual comparison of 3D point projections from a single frame is compared with the renderings from the reconstruction and CT shown in Fig. \ref{fig:depth_estimates}.

\begin{table}[h]
\centering
\caption{Average and standard deviations of the Target Registration Error (TRE) and Point-to-Mesh Distance between CT Segmentation and Depth Estimate Projections.}
\label{tab:depth_projections}
\resizebox{\columnwidth}{!}{
\begin{tabular}{ccc}
\textbf{Reconstruction} & \textbf{TRE (mm)} & \textbf{Point-to-Mesh (mm)} \\ \hline
Baseline & 6.58 $\pm$~2.5 & 0.91 $\pm$~0.4 \\
Predicted Depth Projections & 10.99 $\pm$1.7 & 1.01 $\pm$0.3 \\
\hline
\end{tabular}%
}
\end{table}

The projected points result in a considerable increase in TRE compared to the baseline, while point-to-mesh distance has a slight increment. It is likely that the point-to-mesh distance remains small as the errors in projected points are within the nasal cavity and are still close to the sinus wall. We observe that while fusion seems to refine the depth prediction to produce similar depth renderings of the reconstruction and CT, the projected distances are significantly shorter, which may introduce noise that results in the warping of the overall structure. We then investigated the agreement of reconstructions based on spatially localized endoscopic sequences, where differences between local and global consistency are less significant.

\subsection{Depth Consistency in Local vs. Global Reconstructions} \label{sec:local_reconstruction}

Previous experiments evaluated reconstructions generated with sequences that broadly cover the sinus anatomy. Although these reconstructions produce a more comprehensive representation of the anatomy, the long trajectory of the endoscope might allow for a more significant margin of error in the global consistency of depth predictions. Considering this as a potential source of error in depth estimation, we investigated the performance of the reconstruction pipeline for shorter trajectories, where the differences between long and short depths are less significant.

\begin{figure}[t]
    \centerline{\includegraphics[width=0.9\columnwidth]{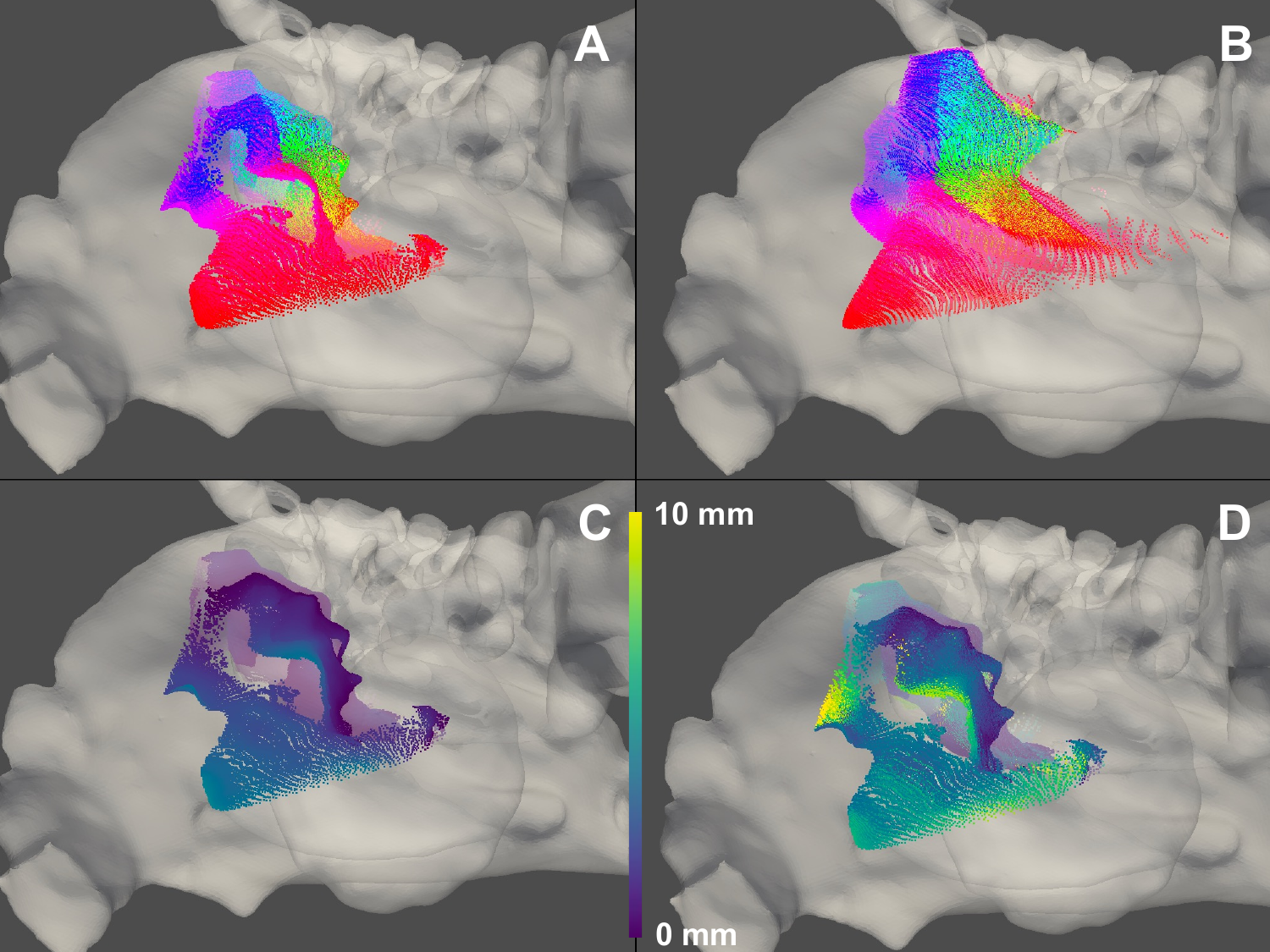}}
    \caption{Top Row: Spatial distribution of the color-coded correspondences employed for TRE computation of the endoscopic sub-sequence for: (A) the dense reconstruction and (B) the CT segmentation where the same color indicates a correspondence. Bottom Row: Spatial distribution of errors of the reconstruction's keypoints for: (C) the point-to-mesh distance and (D) the TRE. Results are presented in contrast with the full CT segmentation (indicated in gray)}
    \label{fig:local_dreco}
\end{figure}
 
To evaluate the impact of global consistency, we visually selected a set of nine sub-sequences (one per endoscopic video). Each sub-sequence focuses on regions around structures close to the anatomy surface (for example, a sequence that inspects the turbinate), leading to spatially-local trajectories. These poses reduce the distributions of distances between the camera and the anatomy surface, reducing the variance in the magnitude of depths. Hence, we expect to have improved consistency in the depth predictions.  
\begin{table}[h]
\centering
\caption{Average and standard deviations of the Target Registration Error (TRE) and Point-to-Mesh Distance between the CT segmentation, the baseline reconstructions, and reconstructions performed with a local trajectory.}
\label{tab:local_seq_results}
\resizebox{\columnwidth}{!}{
\begin{tabular}{ccc}
\hline
\textbf{Reconstruction} & \textbf{TRE (mm)} & \textbf{Point-to-Mesh (mm)} \\ \hline
Baseline & 6.58 $\pm$~2.5 & 0.91 $\pm$~0.4 \\ 
Local & 3.75 $\pm$~1.3 & 0.88 $\pm$~0.5 \\ 
\hline
\end{tabular}%
}
\end{table}

The resulting reconstructions present an overall improvement in the TRE (Table \ref{tab:local_seq_results}), suggesting a more substantial agreement in the 3D correspondences between the depth CT and the reconstruction. This is visually presented in Fig. \ref{fig:local_dreco}, where the distribution of 3D correspondences presents a better agreement across the CT segmentation in Fig. \ref{fig:local_dreco} A and Fig. \ref{fig:local_dreco} B. While the distance distribution of the point-to-mesh metric still shows overall lower values (Fig. \ref{fig:local_dreco} C), the distribution of TRE errors shows a reduction (Fig. \ref{fig:local_dreco} D), indicating an improvement in the local/global agreement in the depth predictions.

\section{Conclusion} \label{sec:conclusion}
 
Our work evaluates the dense reconstruction pipeline presented in \cite{bib:liudreco2020} and investigates the influence of pose and depth estimation accuracy in the volumetric fusion that leads to the final 3D anatomical surface. Employing the point-to-mesh distance and the TRE as main metrics, our findings show that in the baseline implementation, the mapping of reconstructed points does not completely align with the CT (TRE = 6.58\,mm) even though the surfaces may be similar (Point-to-Mesh Error = 0.91\,mm). The reconstructed mesh appears to have local regions with lower TREs and high TREs in others, resulting in warping of the overall structure. Further, after evaluating the effect of pose and depth estimation accuracy, we found that both components contribute equally to the TRE, increasing this error by similar magnitudes with respect to reconstructions based on ideal depths and poses obtained from CT ground-truth. Additionally, we observed that shorter endoscopic sequences reconstructed using a sub-trajectory of the baseline ultimately improved the TRE (3.75\,mm).

We conclude that current efforts should not only focus on accurate depth estimations; it is also necessary to enforce global consistency in the depth estimation to improve agreement between predictions and the ground-truth anatomy. Furthermore, consistent camera pose estimation is equally important, indicating that we should focus on co-optimizing both components to refine the extracted information throughout the reconstruction pipeline. 
By pursuing this goal, we can ensure proper synergy between each element, ultimately resulting in improved reconstructions.
These advancements will provide a strong foundation towards a practical implementation of this innovative technology in clinical settings.

\bibliographystyle{IEEEtran}
\bibliography{references}

\end{document}